\documentclass[article]{IEEEtran}
\usepackage{soul, xcolor}
\sethlcolor{cyan}

\usepackage{pifont}
\usepackage{booktabs}
\usepackage{tabularx}
\usepackage{array}
\usepackage{graphicx} 
\usepackage{lettrine}
\usepackage{multirow}
\usepackage{enumitem}

\usepackage[table]{xcolor}
\definecolor{tableblue}{RGB}{218, 232, 252}
\definecolor{tablegreen}{RGB}{213, 232, 212}
\definecolor{tableyellow}{RGB}{255, 230, 204}
\definecolor{tablered}{RGB}{249, 206, 204}
\definecolor{tablepurple}{RGB}{225, 213, 231}
\begin{document}
\newcolumntype{Y}{>{\raggedright\arraybackslash}X}

\title{A Taxonomy of Cognitive Capability Gaps in Generative and Agentic AI}

\author{\IEEEauthorblockN{Taye Akinrele\IEEEauthorrefmark{1},
Sindhuja Penchala\IEEEauthorrefmark{2},
Noorbakhsh Amiri Golilarz\IEEEauthorrefmark{3},
Sudip Mittal \IEEEauthorrefmark{4}
Shahram Rahimi\IEEEauthorrefmark{5}
}
 
\IEEEauthorblockA{Department of Computer Science, The University of Alabama, \\
Tuscaloosa, Alabama, USA.\\
\{toakinrele\IEEEauthorrefmark{1},
spenchala\IEEEauthorrefmark{2},
namirigolilarz\IEEEauthorrefmark{3},
smittal\IEEEauthorrefmark{4}
srahimi1\IEEEauthorrefmark{5}\}@ua.edu}
}
\maketitle

\begin{abstract}
Cognitive AI seeks to move beyond language generation and autonomous task execution toward systems capable of sustained reasoning, adaptive behavior, persistent memory, and self-regulation. While generative and agentic AI have demonstrated impressive capabilities across a wide range of tasks, many fundamental cognitive functions remain fragmented or weakly developed, limiting reliable operation over extended time horizons. This paper presents a taxonomy-driven survey of the major cognitive capability gaps that continue to constrain the development of Cognitive AI. The literature is organized around five dimensions: persistent state modeling, goal-directed autonomy, self-monitoring and control, environment interaction, and learning and adaptation. For each dimension, we review recent advances, identify recurring limitations, and discuss open research challenges. Building on these insights, we outline a conceptual Adaptive Cognitive Intelligence Architecture (ACIA) and examine emerging directions in cognition-centric evaluation. The proposed taxonomy provides a unified framework for organizing existing research, identifying unresolved challenges, and guiding the design of future cognitively capable systems. Together, the taxonomy, architectural perspective, and evaluation framework offer a roadmap for advancing AI systems that exhibit more reliable long-term reasoning, adaptive decision-making, and continual learning. The survey highlights key research opportunities toward more adaptive, reliable, and cognitively capable AI systems, providing a foundation for future progress toward Cognitive AI and, ultimately, Artificial General Intelligence (AGI).
\end{abstract}


\section{Introduction}

\lettrine{A}{rtificial} intelligence (AI) systems are increasingly being deployed in complex, dynamic, and high-stakes environments, including healthcare, finance, scientific discovery, cybersecurity, and autonomous systems. Recent advances in large language models (LLMs), foundation models, and agentic AI have enabled systems capable of language understanding, reasoning, planning, tool use, and autonomous task execution across a wide range of domains \cite{wang2024knowledge, Plaat_2025}. These developments have created the impression that AI systems are progressing toward increasingly intelligent and autonomous behavior.

However, strong performance on reasoning benchmarks, instruction following, and task execution should not be conflated with cognition. Human cognition extends beyond generating correct outputs and includes persistent memory, adaptive learning, self-monitoring, goal maintenance, environmental grounding, and the ability to continuously revise beliefs and behavior in response to experience. While contemporary generative and agentic systems exhibit impressive task-level capabilities, many of these fundamental cognitive properties remain weakly developed or entirely absent \cite{paisner2014goal, wilie-etal-2024-belief, huang2026failurelatentstatepersistence}.

This distinction becomes increasingly important as AI systems move beyond isolated interactions and are expected to operate autonomously over extended time horizons. Long-term autonomous behavior requires maintaining internal state across interactions, adapting to changing environments, monitoring reasoning processes, and sustaining coherent objectives despite evolving conditions. Yet most modern AI systems remain fundamentally reactive, relying on next-token prediction and static training data rather than continuously evolving internal representations \cite{shi2025continual, salwa2025continual}. Although these systems can solve many short-horizon tasks, they often struggle to maintain persistent memory, preserve long-term goals, reason consistently under changing information, and adapt reliably over time \cite{huang2026failurelatentstatepersistence, chang2025alasstatefulmultillmagent}.

Several limitations contribute to this broader cognitive gap. Current systems exhibit weak persistent memory and limited mechanisms for maintaining coherent state across extended interactions, often requiring information to be repeatedly reintroduced through prompts or external memory systems \cite{wilie-etal-2024-belief}. In addition, attention-control mechanisms remain limited for dynamically prioritizing relevant information across long-horizon tasks. While transformer attention enables context-dependent processing, it does not provide persistent cognitive attention capable of selectively focusing on goals, beliefs, memories, and environmental signals over time. Consequently, important information may be overlooked or inconsistently integrated during reasoning and decision-making \cite{parimi2025adaptation, wu2024continual}.

Reasoning and world modeling also remain brittle. While models can retrieve facts and perform localized reasoning, they frequently fail to propagate updates consistently across related beliefs and internal representations, resulting in logical inconsistencies, unstable world models, and unreliable behavior \cite{luo2025knowledgesmithuncoveringknowledgeupdating, dAliberti2026illusion}. Similarly, self-monitoring and metacognitive capabilities remain limited, reducing the ability of systems to recognize uncertainty, detect reasoning failures, evaluate their own performance, or reliably self-correct \cite{kamoi-etal-2024-llms, yang2026llmsadmitmistakesunderstanding}. Furthermore, many agentic systems exhibit weak environmental grounding and adaptive autonomy, relying heavily on externally defined goals and workflows rather than restructuring internal representations in response to changing conditions \cite{anokhin2025arigraph, arike2025technicalreportevaluatinggoal}.

\begin{figure*}[h]
    \centering
    \includegraphics[width=0.60\linewidth]{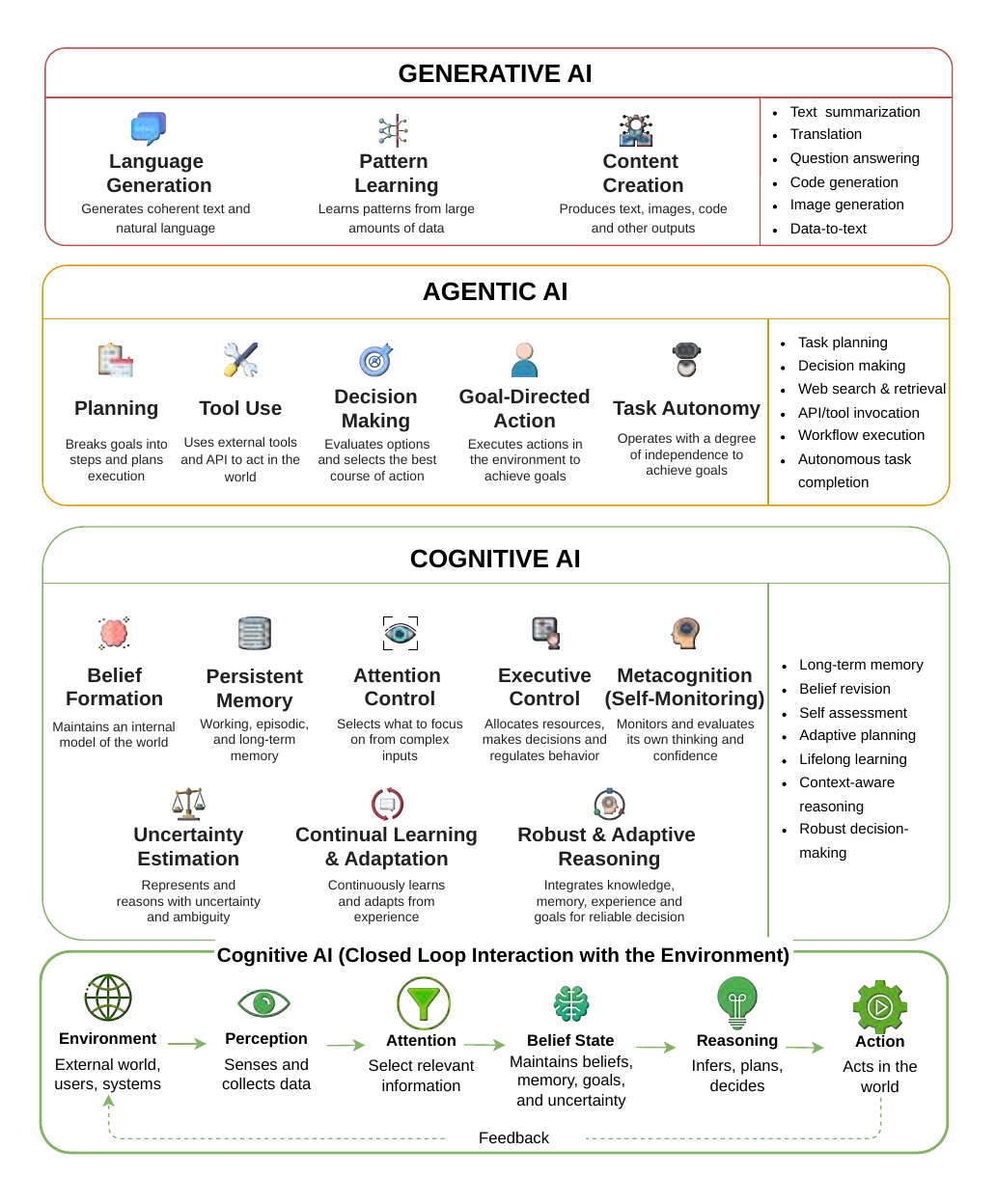}
    \caption{Conceptual evolution from Generative AI to Agentic AI and Cognitive AI, highlighting the additional cognitive capabilities introduced at each stage and the closed-loop perception–reasoning–action cycle underlying sustained cognitive behavior.}
    \label{fig:conceptcognitiveAI}
\end{figure*}

These limitations present significant challenges for real-world deployment. In safety-critical domains, an inability to maintain persistent state, adapt to distribution shifts, revise beliefs, or recognize uncertainty can result in degraded performance and unreliable behavior \cite{ji-etal-2023-towards, arike2025technicalreportevaluatinggoal}. Moreover, continual retraining, knowledge editing, and evolving data streams introduce additional risks of instability, inconsistency, and catastrophic forgetting \cite{liu2025unlockingefficientscalablecontinual, jiang-etal-2024-learning}. As a result, current systems remain heavily dependent on external supervision and human intervention despite increasing levels of operational autonomy.

These challenges have motivated growing interest in cognitive AI, which seeks to move beyond probabilistic generation toward systems capable of sustained reasoning, adaptive learning, persistent memory, self-regulation, and continuous interaction with dynamic environments \cite{huang-etal-2025-alleviating}. Achieving this vision likely requires architectures that integrate memory, reasoning, goal management, metacognitive monitoring, environmental interaction, and adaptive learning within a unified cognitive framework \cite{anokhin2025arigraph, arike2025technicalreportevaluatinggoal}. As illustrated in Figure~\ref{fig:conceptcognitiveAI}, cognitive AI extends beyond content generation and autonomous task execution by maintaining an evolving internal cognitive state composed of beliefs, memory, goals, and self-monitoring mechanisms that support continual adaptation in dynamic environments. While recent surveys have examined important areas such as autonomous agents, continual learning, uncertainty estimation, and agentic AI \cite{shi2025continual, 10.1145/3786319, Plaat_2025}, these efforts are typically organized around individual research domains rather than the broader cognitive capabilities required for robust intelligence. Consequently, a unified understanding of the major cognitive limitations that continue to constrain current AI systems remains underdeveloped.

To address this gap, this paper presents a taxonomy-driven analysis of the core cognitive limitations of modern generative and agentic AI systems and discusses their implications for Cognitive AI. The proposed taxonomy organizes these limitations into five interconnected cognitive components: persistent state modeling, goal-directed autonomy, self-monitoring and control, environment interaction, and learning and adaptation (Fig.~\ref{fig:taxonomy}). These components represent the fundamental capabilities required for sustained cognitive behavior and highlight recurring limitations that continue to constrain the development of more adaptive, autonomous, and cognitively capable AI systems.

The primary contributions of this survey are as follows:

\begin{itemize}
    \item We propose a taxonomy of core cognitive capability gaps in contemporary generative and agentic AI systems, organized around five fundamental dimensions of cognition.
    
    \item We synthesize existing research on memory, reasoning, autonomy, metacognition, environmental grounding, and adaptation, highlighting persistent limitations that remain unresolved.
    
    \item We present a recommended Adaptive Cognitive Intelligence Architecture (ACIA) that integrates memory, reasoning, metacognition, action, and adaptive learning within a unified cognitive framework.
    
    \item We examine current evaluation methodologies and discuss cognition-centric evaluation strategies for assessing persistent memory, adaptive behavior, and cognitive consistency beyond conventional benchmark performance.
    
    \item We identify key research challenges and future directions toward the development of more adaptive, reliable, and cognitively capable AI systems.

\end{itemize}

The remainder of this paper is organized as follows. Section \ref{sec:background} introduces Cognitive AI and its core capability requirements. Section \ref{sec:survey} presents the survey methodology and taxonomy overview. Section \ref{sec:taxonomy} discusses the proposed cognitive capability gaps and associated limitations in current AI systems. Section \ref{sec:architecture} presents the recommended Adaptive Cognitive Intelligence Architecture (ACIA), while Section \ref{sec:evaluation} examines evaluation strategies and challenges. Section \ref{sec:researchchallenge} discusses open research questions and future directions, and Section \ref{sec:conclusion} concludes the paper.

\section{Background}
\label{sec:background}

This section establishes the conceptual foundation for the survey. We first define Cognitive AI and distinguish it from related paradigms, then discuss the evolution of AI systems toward increasingly autonomous and adaptive behavior. Finally, we introduce the core cognitive capability dimensions that motivate the taxonomy used throughout the remainder of this paper.

\begin{figure*}[h]
    \centering
    \includegraphics[width=0.95\linewidth]{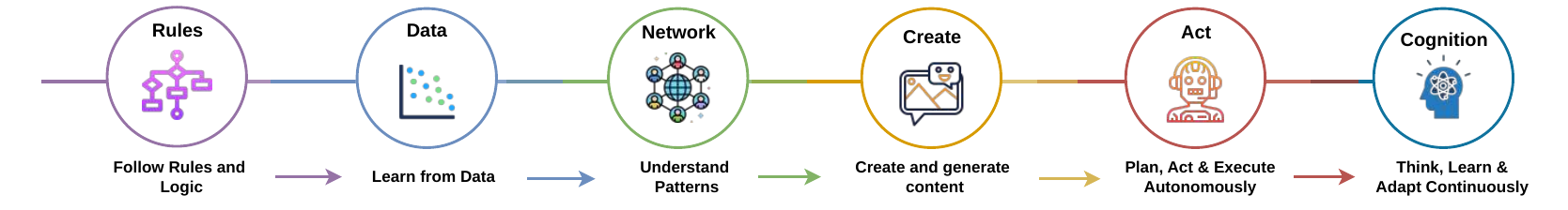}
    \caption{Evolution of artificial intelligence paradigms from rule-based systems to Cognitive AI.}
    \label{fig:evolution}
\end{figure*}

\begin{table*}[!ht]
\caption{Evolution of Artificial Intelligence Toward Cognitive AI}
\label{tab:ai_evolution}
\centering
\footnotesize
\renewcommand{\arraystretch}{1.2}

\begin{tabular}{|p{2.2cm}|p{1.8cm}|p{3.0cm}|p{3.2cm}|p{3.0cm}|}
\hline
\rowcolor{tablered}
\textbf{AI Paradigm} & \textbf{Timeline} & \textbf{Core Function} & \textbf{Potential Applications} & \textbf{Major Limitations} 
\\
\hline

\textbf{Rule-Based AI} \cite{russell2010artificial} & 1950s-1980s &
Follows predefined symbolic rules and logical reasoning. &
Expert systems, rule engines, and decision support systems. &
Cannot learn, brittle behavior, and limited adaptability. \\
\hline

\textbf{Machine Learning} \cite{fradkov2020early} & 1990s-2000s &
Learns statistical patterns from structured data. &
Spam filtering, recommendation systems, fraud detection, and prediction tasks. &
Requires feature engineering and struggles with complex relationships. \\
\hline

\textbf{Deep Learning} \cite{pouyanfar2018survey} & 2010s &
Learn hierarchical representations from large-scale data using neural networks. &
Image recognition, speech processing, machine translation, and autonomous perception. &
High computational cost, large data requirements, and limited interpretability. \\
\hline

\textbf{Generative AI} \cite{bubeck2023paper} & Late 2010s-Early 2020s &
Generates text, images, code, audio, and multimodal content. &
Chatbots, copilots, content generation, and summarization. &
Hallucinations, weak reasoning, and limited contextual understanding. \\

\hline

\textbf{Agentic AI} \cite{10849561} & From 2024  &
Plans, decides, and executes tasks autonomously using tools and APIs. &
AI assistants, workflow orchestration, and task automation. &
Goal misalignment, limited long-term memory, and need for human oversight. \\
\hline

\textbf{Cognitive AI} \cite{golilarz2025bridging} & Emerging &
Integrates reasoning, memory, adaptive learning, and self-monitoring. &
Autonomous systems, scientific discovery, and complex decision-making. &
Challenges in alignment, safety, trust, and integrating cognitive capabilities. \\
\hline

\end{tabular}
\end{table*}

\subsection{Cognitive Artificial Intelligence}

Artificial General Intelligence (AGI) is often described as the ability of an intelligent system to understand, learn, reason, and adapt across a broad range of tasks and environments with a level of flexibility comparable to human intelligence \cite{goertzel2014artificial}. Achieving this objective requires capabilities that extend beyond task-specific performance, including persistent memory, adaptive learning, goal-directed behavior, self-monitoring, and the ability to continuously interact with and learn from dynamic environments. These capabilities are commonly associated with cognition and are widely regarded as fundamental requirements for the development of more general and autonomous forms of intelligence.

Within this context, Cognitive Artificial Intelligence has emerged as a research direction focused on developing systems that exhibit persistent, adaptive, and self-regulating behavior. Cognitive AI seeks to integrate perception, memory, reasoning, learning, action, and metacognitive control within a unified feedback-driven framework \cite{golilarz2025bridging, 10849561}. Unlike systems that treat interactions as isolated events, cognitive systems maintain internal state, accumulate experience over time, revise beliefs in response to new information, and adapt their behavior based on environmental feedback. The objective is to support sustained intelligence capable of operating effectively in complex and evolving environments.

The emergence of generative and agentic AI has brought the field closer to this objective. Generative AI has demonstrated remarkable capabilities in language understanding, content generation, and knowledge synthesis, while agentic AI extends these capabilities through planning, tool use, and autonomous task execution \cite{10849561}. Despite these advances, many existing systems remain dependent on external orchestration and continue to exhibit limitations in memory persistence, long-term adaptation, self-monitoring, and autonomous decision-making \cite{anokhin2025arigraph, huang2026failurelatentstatepersistence}. Cognitive AI seeks to address these limitations by integrating cognitive capabilities within a cohesive architecture that supports continual adaptation, self-regulation, and long-horizon autonomy. As such, Cognitive AI may be viewed as an important step toward the development of more robust, adaptive, and generally intelligent systems.

\subsection{Evolution of Artificial Intelligence to Cognitive Intelligence}

The development of Cognitive AI can be viewed as part of a broader progression in artificial intelligence (see Fig \ref{fig:evolution}). Early AI systems relied primarily on symbolic reasoning and predefined rules to perform deterministic tasks \cite{russell2010artificial}. The emergence of machine learning shifted the focus toward data-driven pattern recognition and statistical inference, while deep learning enabled increasingly sophisticated representation learning from large-scale data \cite{fradkov2020early, pouyanfar2018survey}. More recently, foundation models and generative AI systems demonstrated remarkable capabilities in language understanding, multimodal processing, reasoning, and content generation \cite{bommasani2021opportunities, bubeck2023paper}.

The rise of Agentic AI further expanded these capabilities by incorporating planning, tool use, and autonomous task execution \cite{10849561}. Despite these advances, important challenges related to memory persistence, continual adaptation, self-monitoring, and long-term autonomy remain unresolved \cite{arike2025technicalreportevaluatinggoal, huang2026failurelatentstatepersistence}. Cognitive AI represents the next stage in this evolution by emphasizing persistent cognition rather than short-term task completion. In this paradigm, intelligence emerges from the interaction of memory, reasoning, learning, self-regulation, and environmental feedback rather than from generation alone \cite{golilarz2025bridging}.

Table~\ref{tab:ai_evolution} summarizes the progression of AI paradigms and highlights how Cognitive AI extends previous approaches by integrating capabilities necessary for sustained autonomous behavior.

\section{Survey Methodology and Taxonomy Overview}
\label{sec:survey}

This survey adopts a narrative synthesis approach to examine the cognitive limitations of contemporary generative and agentic AI systems. Rather than organizing the literature by model family or application domain, we analyze existing research through the lens of cognitive capabilities and operational behavior.

The surveyed literature spans cognitive architectures, foundation models, agentic AI, memory systems, reasoning and planning, metacognition, environment interaction, world modeling, and continual learning. Particular attention is given to recurring limitations that affect long-term autonomy, including memory degradation, reasoning inconsistency, goal drift, weak environmental grounding, uncertainty management, and continual adaptation.

Through this synthesis, five recurring dimensions of cognitive capability emerged across the surveyed literature: Persistent State Modeling, Goal-Directed Autonomy, Self-Monitoring and Control, Environment Interaction, and Learning and Adaptation. These dimensions form the taxonomy used throughout the remainder of this survey.

\begin{figure*}[h]
    \centering
    \includegraphics[width=0.80\linewidth]{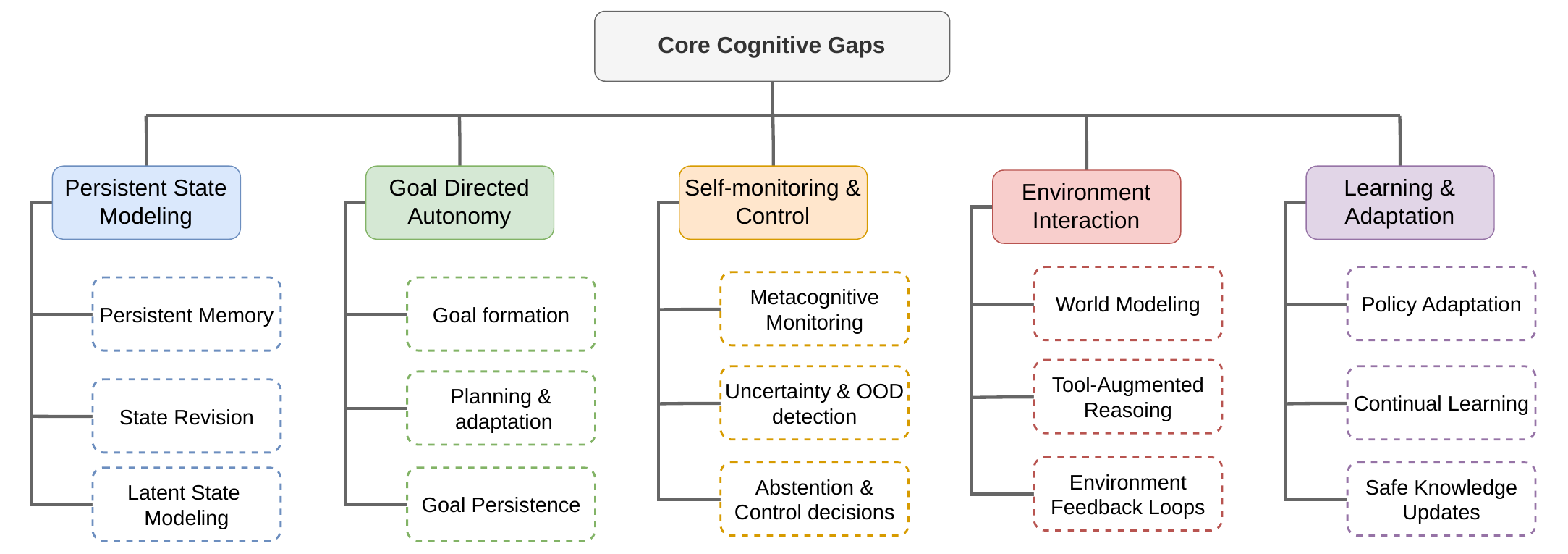}
    \caption{Taxonomy of core cognitive capability gaps in modern AI systems, organized into five cognitive components: persistent state modeling, goal-directed autonomy, self-monitoring and control, environment interaction, and learning and adaptation.}
    \label{fig:taxonomy}
\end{figure*}

\section{Taxonomy of Core Cognitive Capability Gaps}
\label{sec:taxonomy}

This section presents the proposed taxonomy of the core cognitive capability gaps in modern generative and agentic AI systems, as illustrated in Fig.~\ref{fig:taxonomy}. Table~\ref{tab:cognitive_gaps} provides a summary of the proposed taxonomy. The following subsections discuss each cognitive capability component, its associated objectives, limitations, and representative approaches from the literature.

\subsection{Persistent State Modeling}
Similar to human working memory, which supports sophisticated reasoning, persistent state modeling describes an AI system's capacity to preserve, update, and modify internal representations across time \cite{huang2026failurelatentstatepersistence}. While large language models (LLMs) excel at immediate text generation, they typically lack stable internal representations of the world and instead operate as reactive, post hoc solvers rather than proactive planners. This limitation becomes particularly pronounced in long-horizon tasks, where earlier context must be revised or integrated with new information.

This subsection outlines the key limitations of persistent state modeling, along with representative approaches proposed to address them and the gaps that remain.

\subsubsection*{\textbf{Persistent Memory}}
A core requirement of cognitive AI is the ability to retain and integrate information across interactions, tasks, and time horizons, enabling behavior that extends beyond reactive inference \cite{park2023generativeagentsinteractivesimulacra}. However, fixed context windows and stateless interactions often cause prior information to be displaced as new content arrives, limiting the ability of current systems to accumulate experience over time \cite{hu2025memory}.

Recent work has explored a range of approaches to address this limitation. Architectures such as Transformer-XL \cite{dai-etal-2019-transformer} and Compressive Transformers \cite{rae2019compressivetransformerslongrangesequence} improve long-range dependency modeling, while frameworks such as MemGPT \cite{packer2024memgptllmsoperatingsystems} augment LLMs with hierarchical memory management. Other systems, including AriGraph \cite{anokhin2025arigraph}, MemoryBank \cite{zhong2024memorybank}, and SYNAPSE \cite{jiang2026synapseempoweringllmagents}, organize experiences through structured memory representations and associative retrieval mechanisms.

These methods substantially improve memory retention, retrieval accuracy, and long-horizon interaction capabilities. However, memory remains largely external to the model rather than maintained as a continuously evolving cognitive state. The resulting challenges in long-term state management are further exacerbated by the “curse of memory,” where modeling long-term dependencies becomes increasingly costly as task complexity grows \cite{li2024cursememoryrecurrentneural}.

\subsubsection*{\textbf{State Revision}}
Beyond retaining information, cognitive systems must be capable of revising beliefs in response to new or conflicting evidence. While LLMs often appear to update their reasoning during chain-of-thought generation, empirical studies suggest that these revisions are frequently shallow and inconsistent. For example, the Belief-R benchmark demonstrates that many models struggle to reliably update conclusions when presented with new information, often exhibiting trade-offs between adaptability and stability \cite{wilie-etal-2024-belief}.

Current approaches address this challenge through knowledge editing, reflective reasoning, and search-based refinement. Knowledge editing methods such as LTE \cite{jiang-etal-2024-learning} and BaFT \cite{liu2025unlockingefficientscalablecontinual} selectively update stored knowledge, while self-reflection and self-correction methods attempt to refine outputs through iterative feedback \cite{ji-etal-2023-towards}. Search and backtracking approaches further seek to improve reasoning through exploration of alternative solution paths \cite{qin2025backtrackbacktracksequentialsearch}.

These findings point to a deeper structural gap: current systems can retrieve or modify information, but they do not reliably revise beliefs in a coherent and stable manner under uncertainty. In other words, updating outputs or accessing new knowledge does not amount to maintaining a consistent, revisable internal belief state \cite{huang2024largelanguagemodelsselfcorrect, kamoi-etal-2024-llms, yang2026llmsadmitmistakesunderstanding}.


\subsubsection*{\textbf{Latent State Modeling}}
Robust cognition depends not only on explicit reasoning but also on the ability to maintain stable internal representations that preserve commitments, track evolving conditions, and support consistent behavior over time. Current AI systems rely heavily on language-based reasoning, requiring intermediate reasoning states to be expressed explicitly rather than maintained internally \cite{chen2025reasoninglanguagecomprehensivesurvey}. This contributes to the latent state persistence gap, where models fail to maintain stable hidden commitments across related interactions, resulting in inconsistency and concept drift \cite{huang2026failurelatentstatepersistence}.

Recent approaches attempt to shift part of the reasoning process into latent space. Approaches such as Coconut \cite{hao2025traininglargelanguagemodels} perform reasoning through continuous hidden representations, while CTRLS \cite{wu2026ctrlschainofthoughtreasoninglatent} models reasoning as latent state transitions under uncertainty. Related work connecting State Space Models and Hidden Markov Models further highlights the potential of latent-state representations for reasoning and decision-making \cite{ghojogh2026relationstatespacemodels}.

While these approaches expand the representational space available for reasoning, most remain confined to within-sequence inference and do not maintain persistent latent state across interactions. Consequently, current systems continue to reason primarily over generated traces rather than stable internal commitments, limiting their ability to sustain coherent world models over time.

\subsection{Goal Directed Autonomy}
Goal-directed autonomy refers to the ability of an AI system to formulate objectives, plan and adapt actions, and sustain goal pursuit over time. In contrast, most current AI systems operate reactively, responding to prompts without maintaining stable goals across interactions. While recent agentic systems incorporate capabilities such as planning, tool use, and task decomposition \cite{10849561}, they remain dependent on externally specified instructions and lack the ability to autonomously maintain and pursue goals over extended interactions. This limitation becomes particularly evident in multi-step or open-ended settings, where sustained goal commitment is required.

This subsection examines the key limitations of goal-directed autonomy, organized around goal formulation, planning and adaptation, and goal persistence, along with representative approaches and the gaps that remain.

\subsubsection*{\textbf{Goal Formulation}}

A key aspect of autonomy is the ability to generate and refine goals rather than simply executing externally specified instructions. Current AI systems remain predominantly reactive, with behavior driven by prompts, predefined objectives, or reward functions \cite{golilarz2025bridging}. As a result, they struggle to initiate exploration, restructure priorities, or pursue self-directed objectives.

Current methods generally focus on subgoal discovery and hierarchical planning. In reinforcement learning, subgoals can emerge through trajectory analysis, clustering, and anomaly detection, allowing agents to identify intermediate objectives that support learning \cite{rafati2019unsupervised}. Other frameworks employ hierarchical planners and task decomposition mechanisms to bridge high-level goals and executable actions \cite{Webb2025-aj}. While these approaches improve task decomposition and local autonomy, goal formulation remains largely domain-specific and lacks the intrinsic drives required for open-ended exploration, uncertainty reduction, and self-directed learning.

\subsubsection*{\textbf{Planning and Adaptation}}

Effective autonomy requires agents to translate goals into coherent action sequences and adapt those plans as conditions evolve \cite{wei2025plangenllms}. Although modern AI systems can generate plausible plans, they often struggle to maintain consistency, satisfy constraints, and adapt reliably over long horizons \cite{10.1145/3778534.3778661}.

Current approaches generally fall into three categories: reasoning-based planning, hierarchical planning, and adaptive planning. Reasoning-based methods such as Chain-of-Thought \cite{wei2023chainofthoughtpromptingelicitsreasoning}, Tree of Thoughts \cite{yao2023treethoughtsdeliberateproblem}, and Graph of Thoughts \cite{besta2024graph} improve planning through explicit reasoning, exploration, and backtracking. Hierarchical frameworks such as HiPlan \cite{li2025hiplan} support long-horizon planning through multi-level guidance, while adaptive approaches including AdaPlanner \cite{NEURIPS2023_b5c8c1c1}, Reflexion \cite{shinn2023reflexionlanguageagentsverbal}, and Self-Refine \cite{madaan2023selfrefineiterativerefinementselffeedback} incorporate feedback to revise plans and recover from failures.

Despite these advances, planning remains brittle in dynamic environments. Methods that improve adaptability often introduce significant computational overhead, while systems that generate effective plans frequently struggle to preserve consistency across extended reasoning chains. Balancing planning reliability, adaptability, and efficiency therefore remains an open challenge.

\subsubsection*{\textbf{Goal Persistence}}

Beyond generating and executing plans, autonomous systems must maintain commitment to long-term objectives despite environmental changes and competing influences. A common failure mode is goal drift, where an agent gradually deviates from its intended objective as interactions accumulate or conditions evolve \cite{arike2025technicalreportevaluatinggoal}.

Existing work has explored several mechanisms for improving goal stability. Goal-driven autonomy frameworks emphasize sustained goal management rather than isolated task completion \cite{paisner2014goal}, while strong goal elicitation methods seek to suppress competing objectives and reinforce alignment with intended outcomes \cite{arike2025technicalreportevaluatinggoal}. Other approaches include coherence-based alignment \cite{abdicoherence}, embodied autonomy frameworks such as PEPA \cite{liu2026pepapersistentlyautonomousembodied}, and formal measures such as Maximum Entropy Goal-Directedness \cite{NEURIPS2024_1551c01d} that quantify consistency in objective pursuit.

Although these approaches improve goal stability, maintaining long-term goal commitment remains difficult. Agents remain vulnerable to inherited goal drift, environmental distractions, and accumulated errors over extended interactions \cite{menon2026inheritedgoaldriftcontextual}. Consequently, the ability to follow instructions or optimize short-term objectives does not necessarily translate into persistent autonomy over long time horizons.

\subsection{Self-Monitoring and Control}

One of the defining characteristics of cognition is the ability to reflect on and regulate one's own behavior. Effective decision-making requires systems to monitor reasoning processes, estimate uncertainty, identify errors, and adjust actions when conditions change. While recent AI systems exhibit increasingly sophisticated reasoning capabilities, they remain limited in their ability to evaluate the reliability of their own outputs and dynamically regulate behavior without external intervention \cite{10.1145/3613904.3642902, wen2024from}. This subsection examines key limitations in self-monitoring and control and the extent to which current approaches address them.

\subsubsection*{\textbf{Metacognitive Monitoring}}

Robust intelligence requires the ability to evaluate ongoing reasoning, identify potential errors, and revise conclusions when necessary. Current AI systems, however, typically proceed directly to generation without critically assessing intermediate reasoning states, often producing confident outputs despite flawed reasoning \cite{huang2024largelanguagemodelsselfcorrect}. This limitation is particularly problematic in high-stakes settings, where users may interpret fluent responses as evidence of correctness.

Research on metacognitive monitoring has largely focused on three directions: reasoning verification, self-correction, and reflective reasoning. Verification-based methods such as PROCO \cite{wu2024large}, Step CoT Check, and VeriFY \cite{altinisik2026ireallyknowlearning} seek to identify logical inconsistencies in intermediate reasoning steps. Other approaches, including ProgCo \cite{song-etal-2025-progco}, augment linguistic reasoning with symbolic verification, while reflective frameworks attempt to improve outputs through iterative self-evaluation and revision.

While these techniques improve error detection in constrained settings, they fall short of providing robust metacognitive monitoring. Token-level probabilities remain weak indicators of semantic correctness, and models frequently fail to recognize reasoning errors without external guidance. As a result, self-correction remains limited and fragile.

\subsubsection*{\textbf{Uncertainty and Out-of-Distribution Detection}}

Effective self-monitoring also requires the ability to estimate confidence and recognize situations where available knowledge may be insufficient. In practice, AI systems often exhibit overconfidence, particularly on difficult or out-of-distribution (OOD) tasks, and struggle to distinguish uncertainty arising from ambiguous inputs, missing knowledge, or distributional shifts \cite{10.1145/3711896.3736569}.

Efforts to improve uncertainty awareness have centered on confidence estimation, semantic uncertainty, and out-of-distribution detection. Methods such as Answer-Free Confidence Estimation (AFCE) \cite{wen2024from} estimate confidence independently of answer generation, while Semantic Entropy and Kernel Language Entropy measure uncertainty through semantic variation across outputs \cite{10.1145/3711896.3736569}. In multimodal settings, approaches such as EOE leverage generated outlier examples to improve OOD detection \cite{cao2024envisioningoutlierexposurelarge}.

Nevertheless, reliable uncertainty estimation remains elusive across domains and prompting conditions. Small changes in wording can substantially alter confidence estimates without corresponding changes in correctness \cite{he2025out}. More fundamentally, expressions of confidence often reflect learned linguistic patterns rather than calibrated assessments of underlying uncertainty \cite{liu-etal-2025-revisiting}.

\subsubsection*{\textbf{Abstention and Control Decisions}}
An important aspect of cognitive control is recognizing when available evidence is insufficient to support a reliable conclusion. Current AI systems are often optimized to provide answers rather than acknowledge uncertainty, increasing the risk of generating plausible but unsupported responses \cite{wen2025know}. This limitation is compounded by coarse control mechanisms that frequently reduce decisions to either full compliance or complete refusal.

Several strategies have been proposed to improve abstention behavior, including selective prediction, refusal-aware training, and uncertainty-based control mechanisms. Selective prediction methods such as SelectiveNet and RISAN introduce reject options that balance risk and coverage \cite{geifman2019selectivenet, kalra2021risan}. Other approaches, including R-Tuning \cite{zhang2024r}, encourage explicit expressions of uncertainty, while methods such as ABCA \cite{nguyen2025hallucinate} and Seal \cite{huang-etal-2025-alleviating} use knowledge consistency and uncertainty signals to determine when abstention is appropriate.

These strategies reduce unsupported responses but do not fully address the underlying challenge.  Models often struggle to generalize abstention decisions across domains, and efforts to improve refusal behavior can introduce excessive conservatism. More importantly, abstention remains a learned response pattern rather than a grounded decision derived from an explicit understanding of what information is missing and why a reliable answer cannot be provided.

\begin{table*}[!ht]
\centering
\caption{Summary of core cognitive capability gaps in modern generative and agentic AI systems.}
\label{tab:cognitive_gaps}

\footnotesize
\setlength{\tabcolsep}{4pt}
\renewcommand{\arraystretch}{1.25}

\begin{tabular}{|p{2.8cm}|p{2.8cm}|p{3.6cm}|p{6cm}|}
\hline
\rowcolor{tablered}
\textbf{Component} & \textbf{Subcomponent} & \textbf{Objective} & \textbf{Key Limitation} \\
\hline
\multirow{3}{2.8cm}{\textbf{Persistent State Modeling}}
  & Persistent Memory & Maintain long-term state & Memory remains externalized; weakly integrated into core reasoning \\
\cline{2-4}
  & State Revision & Update beliefs consistently & Poor belief propagation; unstable self-correction \\
\cline{2-4}
  & Latent State Modeling & Maintain hidden commitments & No persistent latent state across interactions \\
\hline
\multirow{3}{2.8cm}{\textbf{Goal-Directed Autonomy}}
  & Goal Formulation & Generate internal goals autonomously & No intrinsic drives; goal generation remains domain-specific and non-generalizing \\
\cline{2-4}
  & Planning \& Adaptation  & Decompose and adapt action sequences   & Brittle over long horizons; adaptability gains incur high computational overhead \\
\cline{2-4}
  & Goal Persistence & Sustain objectives over time & Susceptible to goal drift; instruction-following does not guarantee long-term consistency \\
\hline
\multirow{3}{2.8cm}{\textbf{Self-Monitoring \& Control}}
  & Metacognitive Monitoring & Detect and repair reasoning errors & Token probabilities do not reflect semantic correctness; self-correction remains fragile \\
\cline{2-4}
  & Uncertainty \& OOD Detection & Calibrate confidence to correctness & Confidence is prompt-sensitive; linguistic certainty weakly aligned with model probability structure \\
\cline{2-4}
  & Abstention \& Control & Decline unsupported or uncertain claims & Abstention does not generalize across domains; refusal improvements introduce over-abstention \\
\hline
\multirow{3}{2.8cm}{\textbf{Environment Interaction}}
  & World Modeling & Maintain causal environmental models & Poor compositional generalization; systems reason over observations rather than coherent world models \\
\cline{2-4}
  & Tool-Augmented Reasoning & Integrate external tools into reasoning     & Tool use remains loosely coupled to reasoning; models cannot reliably determine when or how to invoke tools \\
\cline{2-4}
  & Environment Feedback Loops   & Incorporate feedback into internal state    & Feedback loops may reinforce bias; online adaptation is difficult to validate in safety-critical settings \\
\hline
\multirow{3}{2.8cm}{\textbf{Learning \& Adaptation}}
  & Policy Adaptation  & Revise decisions under non-stationary conditions  & Adaptation is local and reactive; lacks system-level self-monitoring for persistent, safe policy revision \\
\cline{2-4}
  & Continual Learning & Accumulate knowledge without forgetting & Assumes fixed task boundaries; forward transfer remains limited in open-ended environments \\
\cline{2-4}
  & Safe Knowledge Updates & Update facts without inducing inconsistency & Local edits degrade global coherence; repeated updates increase hallucination risk over time \\
\hline
\end{tabular}
\end{table*}

\subsection{Environment Interaction}

Intelligent behavior depends on the ability to perceive, act within, and adapt to dynamic environments. While modern AI systems can generate plans, invoke tools, and interact with external systems, these capabilities are often only loosely connected to the environments in which actions occur \cite{huang2026failurelatentstatepersistence}. Consequently, many agentic systems operate through predefined workflows and reactive interactions rather than through a grounded understanding of environmental dynamics. This limitation becomes particularly evident in open-ended settings, where successful behavior depends on maintaining coherent representations of the world, integrating external resources into reasoning, and adapting behavior based on feedback.

\subsubsection*{\textbf{World Modeling}}

Effective interaction requires more than observing the environment; it requires maintaining internal models that capture entities, relationships, constraints, and causal dynamics over time. Yet many AI systems continue to rely primarily on statistical associations learned from data rather than explicit models of environmental structure \cite{gupta2025worldmodelsrethinkingunderstanding}. This limits their ability to anticipate consequences, generalize to unfamiliar situations, and adapt when environmental conditions change.

Research in this area has focused on causal world modeling, language-guided environment representations, and structured reasoning frameworks. Approaches that combine causal representation learning with language models seek to capture underlying environmental structure \cite{gkountouras2024languageagentsmeetcausality}, while Language-Guided World Models use language to modify and refine internal dynamics \cite{zhang-etal-2024-language}. Other frameworks, such as CausalARC, employ structural causal models to support abstract reasoning and generalization under limited supervision \cite{maasch2026causalarcabstractreasoningcausal}.

While these methods improve environmental reasoning, robust world modeling remains an open challenge. Many systems continue to struggle with compositional generalization, changing dynamics, and maintaining coherent internal representations over extended interactions. As a result, reasoning often remains grounded in observations rather than in stable, evolving models of the world.

\subsubsection*{\textbf{Tool-Augmented Reasoning}}

Tool use has become a primary mechanism for extending the capabilities of AI systems beyond their parametric knowledge. Modern agents can access search engines, databases, APIs, code interpreters, and robotic interfaces, allowing them to retrieve information and perform actions that would otherwise be impossible. However, access to tools alone does not guarantee effective reasoning. A central challenge is determining when external resources are needed, how retrieved information should influence reasoning, and how actions should be coordinated with ongoing decision-making.

Several frameworks have explored tighter integration between reasoning and tool use. ChatCoT \cite{chen-etal-2023-chatcot} models interaction as a multi-turn reasoning process, while RecThinker \cite{zhang2026recthinkeragenticframeworktoolaugmented} introduces an Analyze--Plan--Act paradigm for more autonomous decision-making. Iterative systems such as DELI \cite{NEURIPS2023_4a47dd69} further support verification and refinement of intermediate reasoning steps.

Nevertheless, tool use remains weakly integrated with cognition. Models frequently struggle to determine when external information is necessary, how tool outputs should update internal reasoning, and when failures should trigger plan revision. Consequently, external capabilities often function as add-on utilities rather than components of a unified cognitive process.

\subsubsection*{\textbf{Environment Feedback Loops}}

Adaptive behavior requires agents to incorporate environmental feedback into future decisions. Although many systems can observe outcomes and react to new information, feedback is often represented as static interaction history rather than as an evolving internal state \cite{Bhat_2025}. This limits the ability of agents to learn from experience, recover from errors, and adapt behavior over extended interactions.

Work in this area has focused on metacognitive feedback mechanisms, reflective architectures, and closed-loop control systems. Metacognitive frameworks enable agents to evaluate and revise their behavior based on observed outcomes \cite{kim2026metacognitive}, while architectures such as BrainBody-LLM introduce hierarchical feedback loops that support error detection and recovery during execution \cite{Bhat_2025}. These approaches move beyond one-pass generation toward more adaptive interaction.

However, reliable feedback-driven adaptation remains difficult to achieve. Online adaptation introduces challenges related to safety, validation, and robustness, particularly in environments where errors can have significant consequences \cite{10.1145/3529836.3529952}. Moreover, feedback loops do not always improve performance and may reinforce biases or amplify errors over time \cite{glickman2025human}. As a result, many systems remain far from the stable, feedback-grounded interaction required for cognitive autonomy.

\subsection{Learning and Adaptation}
Adaptation remains one of the most significant challenges in modern AI systems. While contemporary models can perform effectively on a wide range of tasks, they often struggle to incorporate new experience, revise behavior, and update knowledge after deployment without compromising previously acquired capabilities \cite{cai2025building, shi2025continual}. As a result, many systems remain dependent on retraining or human intervention when operating in changing environments. Consequently, adaptation in current systems is often localized, brittle, and dependent on retraining or external intervention.

\subsubsection*{\textbf{Policy Adaptation}}

Effective adaptation requires agents to modify decision-making strategies as environmental conditions change. However, most AI systems remain optimized for static distributions and struggle to revise policies when reward structures, user preferences, or environmental dynamics evolve \cite{parimi2025adaptation}. This limitation becomes particularly evident in long-term deployments, where behavior must remain effective under continual change.

Research in this area has focused on reinforcement learning, test-time adaptation, and meta-learning. Reinforcement learning methods such as PPO and Deep Q-Learning enable policies to improve through reward-driven interaction, with applications spanning robotics and educational systems \cite{tang2025deep, wu2026proximal}. Test-time adaptation techniques, including AdaContrast, modify representations during inference using self-supervised signals \cite{chen2022contrastive}, while meta-learning seeks model initializations that support rapid adaptation to new tasks with limited data \cite{liureasoning}.

While these techniques improve responsiveness to changing conditions, adaptation is typically treated as a local optimization problem rather than a broader cognitive process. As a result, systems can adjust behavior in the short term but often lack the higher-level monitoring needed to ensure that adaptations remain coherent, stable, and aligned over time.

\subsubsection*{\textbf{Continual Learning}}

A defining characteristic of cognition is the ability to accumulate knowledge over time without losing previously acquired capabilities. Modern AI systems, however, remain vulnerable to catastrophic forgetting, where learning new tasks interferes with existing knowledge and degrades performance on earlier tasks \cite{shi2025continual}. Closely related is the alignment tax, where adaptation to specific domains or objectives can reduce broader reasoning and generalization capabilities \cite{lin2024mitigatingalignmenttaxrlhf}.

Researchers have primarily pursued three strategies to address these challenges: replay-based, regularization-based, and architecture-based methods \cite{salwa2025continual}. Replay strategies such as I-LoRA \cite{ren2024analyzingreducingcatastrophicforgetting} preserve prior knowledge by interleaving historical examples during training. Regularization techniques, including Elastic Weight Consolidation (EWC), constrain updates to parameters important for previous tasks \cite{kirkpatrick2017overcoming}, while approaches such as CURLoRA \cite{fawi2024curlora} seek to balance plasticity and stability through parameter-efficient adaptation mechanisms.

These methods improve retention under controlled conditions but remain limited in open-ended environments where task boundaries are unclear and learning objectives continuously evolve. Moreover, forward transfer remains inconsistent, preventing previously acquired knowledge from reliably accelerating future learning \cite{salwa2025continual}.

\subsubsection*{\textbf{Safe Knowledge Updates}}
Adaptation also requires the ability to update knowledge while preserving consistency across related beliefs and representations. In practice, factual information is embedded within interconnected networks of associations, making localized updates difficult to perform without introducing unintended side effects \cite{cohen2024evaluating}. Consequently, systems may appear to incorporate new information successfully while retaining inconsistencies within their broader knowledge structures.

Model editing and external memory mechanisms have emerged as two major directions for addressing this challenge. Approaches such as ROME \cite{meng2022locating} and MEMIT \cite{meng2023masseditingmemorytransformer} identify and modify parameter regions associated with specific facts, while methods such as AlphaEdit \cite{fang2024alphaedit} seek to reduce interference by constraining updates to minimize disruption of unrelated knowledge. These techniques improve the precision and efficiency of factual modification without requiring full retraining.

Nevertheless, maintaining global consistency after repeated updates remains difficult. Local edits can introduce unintended distortions, alter related knowledge, and gradually degrade the coherence of internal representations \cite{li2023unveiling}. Until update mechanisms can propagate changes consistently across interconnected beliefs, dynamically adapting AI systems will continue to face challenges in maintaining reliable and coherent knowledge over time.

\begin{figure*}[h]
    \centering
    \includegraphics[width=0.75\linewidth]{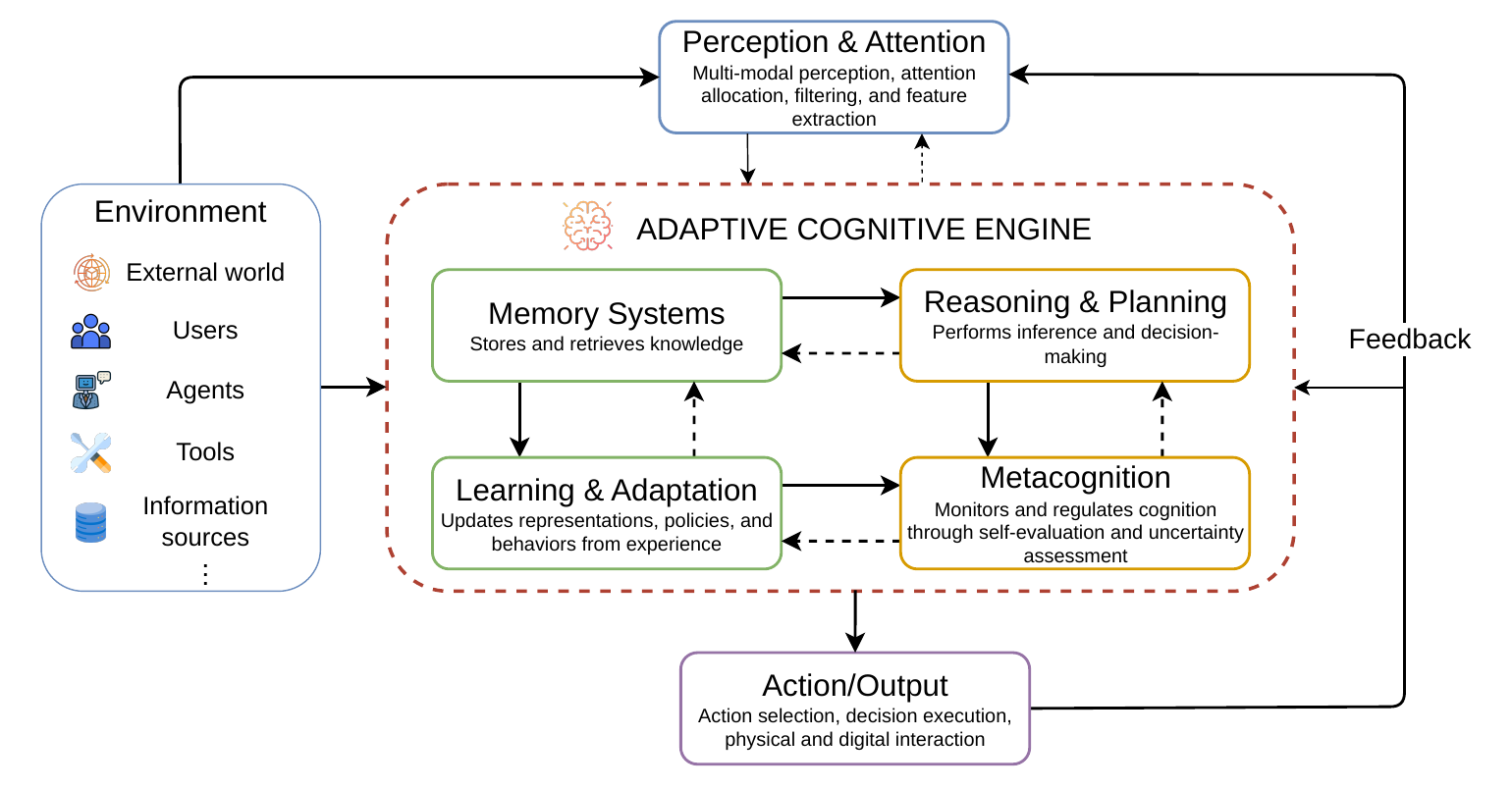}
    \caption{Proposed Adaptive Cognitive Intelligence Architecture (ACIA). The framework models a closed-loop cognitive system consisting of Perception and Attention, Memory, Reasoning and Planning, Metacognition, Action, and Learning and Adaptation}
    \label{fig:architecture}
\end{figure*}

\section{Recommended Architecture for Cognitive AI Systems}
\label{sec:architecture}

\begin{table*}[!ht]
\centering
\caption{Mapping of ACIA components to the cognitive capability gaps identified in this survey.}
\label{tab:acia_mapping}
\footnotesize
\renewcommand{\arraystretch}{1.2}
\setlength{\tabcolsep}{5pt}

\begin{tabular}{|p{2.8cm}|p{3.5cm}|p{3.2cm}|p{4.8cm}|}
\hline
\rowcolor{tablered}
\textbf{Component} & \textbf{Function} & \textbf{Cognitive Domain} & \textbf{Gap Addressed} \\
\hline

Perception \& Attention &
Environmental awareness and context prioritization &
Environment Interaction &
Grounding, world modeling, contextual integration \\
\hline

Memory &
Maintains persistent state and contextual knowledge &
Persistent State Modeling &
Persistent memory, state revision, latent state persistence \\
\hline

Reasoning \& Planning &
Goal-oriented decision making and strategic planning &
Goal-Directed Autonomy &
Goal formulation, planning and adaptation, goal persistence \\
\hline

Metacognition &
Self-monitoring, uncertainty assessment, and control &
Self-Monitoring \& Control &
Metacognitive monitoring, uncertainty estimation, abstention and control \\
\hline

Action &
External execution and interaction with tools and systems &
Environment Interaction &
Tool-augmented reasoning, feedback loops, environment engagement \\
\hline

Learning \& Adaptation &
Continuous improvement through feedback and experience &
Learning \& Adaptation &
Policy adaptation, continual learning, safe knowledge updates \\
\hline

\end{tabular}
\end{table*}

Over the past four decades, cognitive architectures such as ACT-R, Soar, CLARION, LIDA, EPIC, and ICARUS have provided important foundations for modeling reasoning, memory, learning, perception, and autonomous behavior \cite{kotseruba202040}. While modern generative and agentic AI systems demonstrate strong capabilities in language understanding and task execution, many of the cognitive limitations identified throughout this survey remain unresolved.In this section, we propose an Adaptive Cognitive Intelligence Architecture (ACIA) (Fig.~\ref{fig:architecture}), whose components and their relationship to the identified cognitive capability gaps are summarized in Table~\ref{tab:acia_mapping}.

The proposed architecture is organized as a closed-loop cognitive system composed of \textit{Perception and Attention}, \textit{Memory}, \textit{Reasoning and Planning},  \textit{Metacognition}, \textit{Action}, and \textit{Learning and Adaptation}. Together, these components enable continuous interaction between the system and its environment through perception, reasoning, action, feedback, and adaptation.

\begin{itemize}[leftmargin=*]

\item \textbf{\textit{Perception and Attention}}:
Perception and Attention serves as the interface between the architecture and its external environment. It processes multimodal inputs from users, tools, agents, and information sources while prioritizing relevant information and filtering irrelevant stimuli. By maintaining situational awareness and selectively allocating computational resources, this component supports grounded interaction with dynamic environments. It directly addresses limitations in environment interaction by enabling the system to continuously perceive and respond to changing external conditions \cite{golilarz2025bridging, paisner2014goal, jiang2026synapseempoweringllmagents}.

\item \textbf{\textit{Memory}}:
Memory maintains and organizes information required for reasoning, planning, and decision-making. The architecture incorporates working memory for short-term task execution, episodic memory for storing experiences and interactions, and semantic memory for long-term conceptual knowledge. By preserving information across interactions and supporting retrieval, revision, and contextual grounding, memory provides the persistent internal state required for long-horizon cognition. This component directly addresses the persistent state modeling limitations identified in current AI systems \cite{cai2025building, liu2023thinkinmemoryrecallingpostthinkingenable}.

\item \textbf{\textit{Reasoning and Planning}}:
Reasoning and Planning is responsible for inference, problem-solving, decision-making, and goal-directed behavior. It integrates environmental observations with information retrieved from memory to generate plans, evaluate alternatives, and adapt actions as conditions change. Through this process, the architecture can sustain objectives over extended interactions rather than reacting solely to immediate prompts. This component addresses limitations in goal-directed autonomy, particularly in planning, adaptation, and long-term goal maintenance \cite{chang2025alasstatefulmultillmagent, kamoi-etal-2024-llms}.

\item \textbf{\textit{Metacognition}}:
Metacognition provides self-monitoring and self-regulation capabilities. It evaluates confidence, consistency, uncertainty, and reasoning quality while monitoring the effectiveness of ongoing cognitive processes. When errors, contradictions, or uncertainty are detected, metacognitive control can trigger re-planning, memory revision, or additional information gathering. This component addresses the self-monitoring and control limitations discussed throughout the survey \cite{kamoi-etal-2024-llms, yang2026llmsadmitmistakesunderstanding, dAliberti2026illusion}.

\item \textbf{\textit{Action}}:
Action translates cognitive decisions into executable behavior. This includes generating responses, invoking tools, interacting with external systems, coordinating with other agents, and performing autonomous actions. Through continuous interaction with the environment, action closes the perception–reasoning–execution loop and enables the architecture to operate beyond passive generation \cite{cai2025building, arike2025technicalreportevaluatinggoal}.

\item \textbf{\textit{Learning and Adaptation}}:
Learning and Adaptation enables the architecture to improve through experience. Feedback from actions, environmental responses, and task outcomes is used to refine internal representations, update policies, and improve future decision-making. This component supports continual learning, safe knowledge revision, and adaptation to changing environments, addressing the learning and adaptation limitations identified in current AI systems \cite{wu2024large, parimi2025adaptation, wilie-etal-2024-belief}.

\end{itemize}

Unlike contemporary generative and agentic AI systems, which often rely on loosely connected memory, planning, and tool-use mechanisms, ACIA integrates these capabilities within a unified cognitive loop. Persistent memory supports state continuity, reasoning and planning enable goal-directed behavior, metacognition provides self-monitoring and control, and learning mechanisms facilitate continual adaptation. These components address the five cognitive capability gaps identified throughout this survey and provide a conceptual foundation for more adaptive, autonomous, and cognitively capable AI systems.

\begin{table*}[!ht]
\centering
\caption{Illustrative cognition-centric evaluation metrics for assessing long-term cognitive behavior.}
\label{tab:cognitive_metrics}
\footnotesize
\renewcommand{\arraystretch}{1.2}
\setlength{\tabcolsep}{5pt}
\begin{tabular}{|p{2.8cm}|p{3.5cm}|p{3.2cm}|p{4.8cm}|}
\hline
\rowcolor{tablered}
\textbf{Metric} & \textbf{Evaluation Objective} & \textbf{Illustrative Formula} & \textbf{Notation} \\
\hline
Cognitive Persistence Index (CPI) &
Does the agent retain and accurately retrieve relevant information across extended interactions? &
$\mathrm{CPI}=f\!\left(T_{\mathrm{ret}},P_{\mathrm{mem}},R_{\mathrm{mem}}\right)$ &
$T_{\mathrm{ret}}$: temporal memory retention rate; $P_{\mathrm{mem}}$: memory precision; $R_{\mathrm{mem}}$: memory recall \\
\hline
Cognitive Adaptation Rate (CAR) &
Does the agent successfully adapt to new information while preserving prior knowledge? &
$\mathrm{CAR}=f\!\left(A_{\mathrm{succ}},A_{\mathrm{opp}}\right)$ &
$A_{\mathrm{succ}}$: adaptation success rate; $A_{\mathrm{opp}}$: adaptation opportunity coverage \\
\hline
Cognitive Consistency Score (CCS) &
Does the agent maintain coherent beliefs, goals, and reasoning over time? &
$\mathrm{CCS}=f\!\left(1-\frac{N_{\mathrm{contra}}}{N_{\mathrm{reason}}}\right)$ &
$N_{\mathrm{contra}}$: number of contradictions; $N_{\mathrm{reason}}$: number of reasoning episodes \\
\hline
\end{tabular}
\end{table*}

\section{Evaluation Strategies and Challenges}
\label{sec:evaluation}

\subsection{Existing Evaluation Strategies}

The methodologies used to evaluate AI systems have evolved alongside advances in model capabilities. Human evaluation remains one of the most widely adopted approaches for assessing qualities such as coherence, relevance, helpfulness, and alignment with human expectations \cite{ouyang2022training}. Benchmark-based evaluation provides a complementary approach by enabling standardized comparison across models. Frameworks such as BIG-bench \cite{srivastava2023beyond} and HELM \cite{liang2022holistic} evaluate capabilities including language understanding, reasoning, knowledge recall, robustness, and safety across diverse tasks and domains.

As AI systems have become increasingly autonomous, specialized benchmarks have emerged to assess capabilities beyond traditional language generation. These include evaluations of belief revision, planning, tool use, adaptation, and autonomous task execution \cite{wilie-etal-2024-belief, 10849561}. To improve scalability, LLM-as-a-Judge paradigms use large language models to evaluate generated responses through ranking and preference modeling, providing an alternative to costly human assessment \cite{zheng2023judging}. Additional evaluation methodologies focus on retrieval and grounding, measuring factual consistency and hallucination behavior \cite{ji2023survey, lin2022truthfulqa}, while red-teaming and adversarial testing assess robustness, safety, and failure modes under challenging conditions \cite{perez2022red, liang2022holistic}.

Collectively, these approaches provide valuable insights into model performance, reliability, and safety. However, they remain largely focused on outputs, task success, and short-horizon behavior rather than the cognitive processes required for long-term autonomous operation.

\subsection{Challenges in Evaluating Cognitive AI}

Evaluating Cognitive AI extends beyond traditional measures of accuracy and task completion because many cognitive capabilities emerge over extended interactions rather than within isolated tasks. This challenge is particularly evident across the five cognitive capability dimensions discussed throughout this survey.

Persistent state modeling requires assessing whether a system can retain, update, and consistently utilize information over time. However, many contemporary models exhibit limitations in memory persistence and latent state maintenance that are often overlooked by benchmarks operating within a single context window \cite{huang2026failurelatentstatepersistence, hu2025memory}. Similarly, goal-directed autonomy is difficult to evaluate because failures such as goal drift emerge gradually during long-horizon operation and may not be apparent in short-term tasks \cite{arike2025technicalreportevaluatinggoal, paisner2014goal}.

Self-monitoring and control present additional challenges because internal reasoning processes are difficult to observe directly. Apparent self-correction may reflect unstable inference rather than genuine metacognitive reasoning, while uncertainty estimation and confidence calibration remain inconsistent across domains \cite{dAliberti2026illusion, kamoi-etal-2024-llms, yang2026llmsadmitmistakesunderstanding}. Environment interaction is similarly difficult to assess using static benchmarks, as grounded behavior depends on maintaining coherent world models and adapting to environmental feedback over time \cite{anokhin2025arigraph}.

Learning and adaptation introduce further complexity. Continual learning systems must incorporate new information while preserving previously acquired knowledge, yet knowledge updates often introduce unintended side effects and inconsistencies that are rarely captured by conventional evaluation methodologies \cite{liu2025unlockingefficientscalablecontinual, jiang-etal-2024-learning}.

Taken together, these challenges highlight a fundamental limitation of existing evaluation frameworks: they primarily assess what a system can do at a particular moment, whereas Cognitive AI requires evaluating the consistency of cognition over time, including whether systems can maintain memory, preserve goals, adapt to new information, and regulate behavior across extended periods of operation.

\subsection{Toward Cognition-Centric Evaluation}

A cognition-centric evaluation framework should align with the five cognitive capability dimensions discussed throughout this survey. This includes evaluating memory retention and belief revision for persistent state modeling, goal persistence and planning stability for autonomy, uncertainty calibration and abstention behavior for self-monitoring, grounding and feedback utilization for environment interaction, and continual learning and knowledge update consistency for adaptation.

Emerging directions such as belief revision benchmarks, latent state persistence tasks, uncertainty-aware reasoning frameworks, and long-horizon agent evaluations provide promising foundations for such assessments \cite{wilie-etal-2024-belief, huang2026failurelatentstatepersistence, wu2026ctrlschainofthoughtreasoninglatent}. More broadly, future evaluation frameworks should emphasize longitudinal assessment, dynamic environments, and adaptive behavior. For example, cognition-centric metrics may evaluate memory retention across extended interactions, the ability to maintain                                                               objectives under changing conditions, and the consistency of belief updates throughout long-horizon reasoning. Table~\ref{tab:cognitive_metrics} presents illustrative cognition-centric evaluation metrics for long-term cognitive capabilities. Each metric is expressed as a function of lower-level measures related to memory, adaptation, or reasoning consistency, where \textbf{$f(\cdot)$} indicates that the specific formulation may vary across applications and evaluation protocols. These metrics provide a broader assessment of whether AI systems can retain knowledge, adapt to changing information, and maintain coherent reasoning over time, complementing conventional task-level benchmarks.

\section{Research Challenges and Future Directions}
\label{sec:researchchallenge}

The cognitive capability gaps identified throughout this survey suggest that progress toward Cognitive AI will require advances beyond scaling model parameters or improving benchmark performance. Although modern generative and agentic systems demonstrate strong capabilities in language understanding, reasoning, and task execution, they continue to exhibit limitations in persistent state modeling, goal-directed autonomy, self-monitoring and control, environmental interaction, and continual adaptation. A central challenge is the development of persistent cognitive state. Current systems primarily rely on finite context windows or external retrieval mechanisms that provide access to historical information without maintaining stable, evolving internal representations. Future research should investigate architectures capable of maintaining persistent memory, revising beliefs in response to new evidence, and constructing grounded world models that evolve continuously through interaction. Another important direction concerns autonomous reasoning and metacognitive control. Beyond generating plans or actions, cognitive systems must be able to monitor their own reasoning, estimate uncertainty, recognize failures, and revise decisions when necessary. Developing reliable mechanisms for self-monitoring, uncertainty calibration, and adaptive control remains essential for improving robustness in long-horizon tasks.

Continual learning and adaptation remain equally important challenges. Future Cognitive AI systems must integrate new knowledge while preserving previously acquired information and maintaining behavioral consistency over time. Addressing catastrophic forgetting, unstable belief updates, and conflicting knowledge representations will be critical for enabling lifelong learning. Progress will also require evaluation methodologies that measure cognitive capabilities directly rather than relying primarily on task-level performance. Emerging benchmarks for belief revision, latent state persistence, uncertainty-aware reasoning, and long-horizon agents represent promising first steps, but broader evaluation frameworks are needed to assess persistent memory, adaptation, reasoning consistency, and goal stability across extended interactions and dynamic environments. Finally, increasing levels of autonomy place greater emphasis on safety, transparency, and human oversight. Future cognitive systems must balance adaptive decision making with interpretability, controllability, and alignment, ensuring that autonomous behavior remains reliable while operating in open-ended environments.

Addressing these challenges will require closer integration of advances in artificial intelligence, cognitive science, neuroscience, and related disciplines. Progress toward Cognitive AI is therefore likely to depend not only on increasingly capable foundation models, but also on principled cognitive architectures that support persistent state, adaptive learning, metacognitive control, and robust long-term behavior.

\section{Conclusion}
\label{sec:conclusion}

This paper examined the major cognitive capability gaps that continue to limit contemporary generative and agentic AI systems, including persistent state modeling, goal-directed autonomy, self-monitoring and control, environment interaction, and learning and adaptation. Through a taxonomy-driven review, we synthesized recent advances, identified recurring limitations, and discussed their implications for the development of Cognitive AI. We further outlined a conceptual Adaptive Cognitive Intelligence Architecture (ACIA) and highlighted the need for cognition-centric evaluation strategies that extend beyond conventional benchmark performance. These perspectives suggest that achieving Cognitive AI will require moving beyond task-oriented intelligence toward systems capable of sustained cognition, adaptive autonomy, and continual learning. By identifying the major cognitive capability gaps and organizing existing research through a unified taxonomy, this survey provides a foundation for future research on cognitively capable AI systems.


{\scriptsize
\bibliographystyle{IEEEtran}
\bibliography{references}
}

\end{document}